\documentclass[11pt,a4paper]{article}
\usepackage[hyperref]{emnlp2020}
\usepackage{times}
\usepackage{latexsym}

\usepackage{microtype}
\usepackage{graphicx}
\usepackage{booktabs}
\usepackage{adjustbox}
\usepackage{amssymb}
\usepackage{wrapfig}

\aclfinalcopy 
\def\aclpaperid{137} 

\title{The Tatoeba Translation Challenge -- Realistic Data Sets for Low Resource and Multilingual MT}

\author{J\"org Tiedemann \\
University of Helsinki \\
\texttt{jorg.tiedemann@helsinki.fi}\\
\url{https://github.com/Helsinki-NLP/Tatoeba-Challenge}}

\date{}

\begin{document}
\maketitle
\begin{abstract}
This paper describes the development of a new benchmark for machine translation that provides training and test data for thousands of language pairs covering over 500 languages and tools for creating state-of-the-art translation models from that collection. The main goal is to trigger the development of open translation tools and models with a much broader coverage of the World's languages. Using the package it is possible to work on realistic low-resource scenarios avoiding artificially reduced setups that are common when demonstrating zero-shot or few-shot learning. For the first time, this package provides a comprehensive collection of diverse data sets in hundreds of languages with systematic language and script annotation and data splits to extend the narrow coverage of existing benchmarks. Together with the data release, we also provide a growing number of pre-trained baseline models for individual language pairs and selected language groups.
\end{abstract}

\section{Introduction}

The Tatoeba translation challenge includes shuffled training data taken from OPUS,\footnote{http://opus.nlpl.eu/} an open collection of parallel corpora \cite{TIEDEMANN12.463}, and test data from Tatoeba,\footnote{https://tatoeba.org/} a crowd-sourced collection of user-provided translations in a large number of languages. All data sets are labeled with ISO-639-3 language codes using macro-languages in case when available. Naturally, training data do not include sentences from Tatoeba and neither from the popular WMT testsets to allow a fair comparison to other models that have been evaluated using those data sets.

Here, we propose an open challenge and the idea is to encourage people to develop machine translation in real-world cases for many languages. The most important point is to get away from artificial setups that only simulate low-resource scenarios or zero-shot translations. A lot of research is tested with multi-parallel data sets and high resource languages using data sets such as WIT$^3$ \cite{cettoloEtAl:EAMT2012} or Europarl \cite{koehn2005europarl} simply reducing or taking away one language pair for arguing about the capabilities of learning translation with little or without explicit training data for the language pair in question (see, e.g., \citet{firat-etal-2016-multi,firat-etal-2016-zero,ha2016multilingual,Lakev-etal-2018}). Such a setup is, however, not realistic and most probably over-estimates the ability of transfer learning making claims that do not necessarily carry over towards real-world tasks.

In the set we provide here we, instead, include all available data from the collection without removing anything. In this way, the data refers to a diverse and skewed collection, which reflects the real situation we need to work with and many low-resource languages are only represented by noisy or very unrelated training data. Zero-shot scenarios are only tested if no data is available in any of the sub-corpora. More details about the data compilation and releases will be given below.

Tatoeba is, admittedly, a rather easy test set in general but it includes a wide variety of languages and makes it easy to get started with rather encouraging results even for lesser resourced languages. The release also includes medium and high resource settings and allows a wide range of experiments with all supported language pairs including studies of transfer learning and pivot-based methods.

\section{Data releases}

The current release includes over 500GB of compressed data for 2,961 language pairs covering 555 languages. The data sets are released per language pair with the following structure, using deu-eng as an example (see Figure~\ref{fig:data-structure}).

\begin{figure}[ht]
{\small
    \begin{verbatim}
data/deu-eng/
data/deu-eng/train.src.gz
data/deu-eng/train.trg.gz
data/deu-eng/train.id.gz
data/deu-eng/dev.id
data/deu-eng/dev.src
data/deu-eng/dev.trg
data/deu-eng/test.src
data/deu-eng/test.trg
data/deu-eng/test.id
    \end{verbatim}}
    \caption{Released data packages: training data, development data and test data. Language labels are stored in ID files that also contain the name of the source corpus for the training data sets.}
    \label{fig:data-structure}
\end{figure}

Files with the extension $.src$ refer to sentences in the source language ($deu$ in this case) and files with extension $.trg$ contain sentences in the target language ($eng$ here). File with extension $.id$ include the ISO-639-3 language labels with possibly extensions about the orthographic script (more information below). In the $.id$ file for the training data there are also labels for the OPUS corpus the sentences come from. We include the entire collection available from OPUS with data from the following corpora: ada83, Bianet, bible-uedin, Books, CAPES, DGT, DOGC, ECB, EhuHac, EiTB-ParCC, Elhuyar, EMEA, EUbookshop, EUconst, Europarl, Finlex, fiskmo, giga-fren, GlobalVoices, GNOME, hrenWaC, infopankki, JRC-Acquis, JW300, KDE4, KDEdoc, komi, MBS, memat, MontenegrinSubs, MultiParaCrawl, MultiUN, News-Commentary, OfisPublik, OpenOffice, OpenSubtitles, ParaCrawl, PHP, QED, RF, sardware, SciELO, SETIMES, SPC, Tanzil, TED2013, TedTalks, TEP, TildeMODEL, Ubuntu, UN, UNPC, wikimedia, Wikipedia, WikiSource, XhosaNavy.

The data sets are compiled from the pre-aligned bitexts but further cleaned in various ways. First of all, we remove non-printable characters and strings that violate Unicode encoding principles using regular expressions and a recoding trick using the forced encoding mode of $recode$ (v3.7), a popular character conversion tool.\footnote{https://github.com/pinard/Recode} Furthermore, we also de-escape special characters (like '\&' encoded as '\&amp;') that may appear in some of the corpora. For that, we apply the tools from Moses \cite{koehn-etal-2007-moses}. Finally, we also apply automatic language identification to remove additional noise from the data. We use the compact language detect library (CLD2) through its Python bindings\footnote{https://pypi.org/project/pycld2/} and a Python library for converting between different ISO-639 standards.\footnote{https://pypi.org/project/iso-639/} CLD2 supports 172 languages and we use the options for "best effort" and apply the assumed language from the original data as the "hint language code". For unsupported languages, we remove all examples that are detected to be English as this is a common problem in some corpora where English texts appear in various places (e.g. untranslated text in localization data of community efforts). In all cases, we only rely on the detected language if it is flagged as reliable by the software.

All corpus data and sub-languages are merged and shuffled using terashuf\footnote{https://github.com/alexandres/terashuf} that is capable to efficiently shuffle large data sets. But we keep track of the original data set and provide labels to recognize the origin. In this way, it is possible to restrict training to specific subsets of the data to improve domain match or to reduce noise. The entire procedure of compiling the Tatoeba Challenge data sets is available from the project repository at \url{https://github.com/Helsinki-NLP/Tatoeba-Challenge}.

The largest data set (English-French) contains over 180 million aligned sentence pairs and 173 language pairs are covered by over 10 million sentence pairs in our collection. Altogether, there are almost bilingual 3,000 data sets and we plan regular updates to improve the coverage. Below, we give some more details about the language labels, test sets and monolingual data sets that we include in the package as well.


\subsection{Language labels and scripts}

We label all data sets with standardized language codes using three-letter codes from ISO-639-3. The labels are converted from the original OPUS language IDs (which roughly follow ISO-639-1 codes but also include various non-standard IDs) and information about the writing system (or script) is automatically assigned using Unicode regular expressions and counting letters from specific script character properties. For the scripts we use four-letter codes from ISO-15924 and attach them to the three-letter language codes defined in ISO-639-3. Only the most frequently present script in a string is shown. Mixed content may appear but is not marked specifically. Note that the code Zyyy refers to common characters that cannot be used to distinguish scripts. The information about the script is not added if there is only one script in that language and no other scripts are detected in any of the strings. If there is a default script among several alternatives then this particular script is not shown either. Note that the assignment is done fully automatically and no corrections have been made. Three example label sets are given below using the macro-languages Chinese (zho), Serbo-Croatian (hbs) and Japanese (jpn) that can use character from different scripts:

{\small
\begin{description}
\itemsep0pt
    \item[Chinese: ] cjy\_Hans, cjy\_Hant, cmn, cmn\_Bopo, cmn\_Hans, cmn\_Hant, cmn\_Latn, gan, lzh, lzh\_Bopo, lzh\_Hang, lzh\_Hani, lzh\_Hans, lzh\_Hira, lzh\_Kana, lzh\_Yiii, nan\_Hani, nan\_Latn, wuu, wuu\_Bopo, wuu\_Hang, wuu\_Hani, wuu\_Hira, yue\_Hans, yue\_Hant, yue\_Latn
    \item[Japanese:] jpn, jpn\_Hani, jpn\_Hira, jpn\_Kana, jpn\_Latn
    \item[Serbo-Croatian: ] bos\_Latn, hrv, srp\_Cyrl, srp\_Latn
\end{description}}

This demonstrates that a data set may include examples from various sub-languages if they exist (e.g. Bosnian, Croatian and Serbian in the Serbo-Croatian case) or language IDs with script extensions that show the dominating script in the corresponding string (e.g. Cyrl for Cyrillic or Latn for Latin script). Those labels can be used to separate the data sets, to test sub-languages or specific scripts only or to remove some noise (like the examples that are tagged with the Latin script (Latn) in the Japanese data set. Note that script detection can also fail in which the corresponding code is missing or potentially wrong. For example, the detection of traditional (Hant) och simplified Chinese (Hans) can be ambiguous and encoding noise can have an effect on the detection.

We also release the tools that we developed for converting and standardizing OPUS IDs and also the tools that detect scripts and variants of writing systems. The package is available from github\footnote{https://github.com/Helsinki-NLP/LanguageCodes} and can be installed from CPAN.\footnote{https://metacpan.org/pod/ISO::639::3 and https://metacpan.org/pod/ISO::639::5} 


\subsection{Multiple reference translations}

Test and development data are taken from a shuffled version of Tatoeba. All translation alternatives are included in the data set to obtain the best coverage of languages in the collection. Development and test sets are disjoint in the sense that they do not include identical source-target language sentence pairs. However, there can be identical source sentences or identical target sentences in both sets, which are not linked to the same translations. Similarly, there can be identical source or target sentences in one of the sets, for example the test set, with different translations. In Figure~\ref{fig:test-data}, you can see examples from the Esperanto-Ladino test set.


\begin{figure}[ht]
{\small
\begin{tabular}{ll}
epo & lad\_Latn\\
\hline
Ĉu vi estas en Berlino? &Estash en Berlin?\\
Ĉu vi estas en Berlino? &Vos estash en Berlin?\\
Ĉu vi estas en Berlino? &Vozotras estash en Berlin?\\
La hundo estas nigra.   &El perro es preto.\\
La hundo nigras.        &El perro es preto.\\
\end{tabular}}
    \caption{Examples of test sentences with multiple reference translations taken from the Esperanto-Ladino test set.}
   \label{fig:test-data}
\end{figure}

The test data could have been organized as multi-reference data sets but this would require to provide different sets in both translation directions. Removing alternative translations is also not a good option as this would take away a lot of relevant data. Hence, we decided to provide the data sets as they are, which implicitly creates multi-reference test sets but with the wrong normalization.

\subsection{Monolingual data}

In addition to the parallel data sets we also provide monolingual data that can be used for unsupervised methods or data augmentation approaches such as back-translation. For that purpose, we extract public data from Wikimedia including source from Wikpedia, Wikibooks, Wikinews, Wikiquote and Wikisource. We extract sentences from data dumps provided in JSON format\footnote{https://dumps.wikimedia.org/other/cirrussearch/current} and process them with jq,\footnote{https://stedolan.github.io/jq/} a lightweight JSON processing tool. We apply the same cleaning steps as we do for the OPUS bitexts including language identification and convert language IDs to ISO-639-3 as before. Sentence boundaries are detected using UDPipe \cite{straka-etal-2016-udpipe} with models trained on universal dependency treebanks v 2.4 and the Moses sentence splitter with language-specific non-breaking prefixes if available. We preserve document boundaries and do not shuffle the data to enable experiments with discourse-aware models. The data sets are released along with the rest of the Tatoeba challenge data.

\section{The translation challenge}

The main challenge is to develop translation models and to test them with the given test data from Tatoeba. The focus is on low-resource languages and to push their coverage and translation quality. Resources for high-resource are also provided and can be used as well for translation modeling of those languages and for knowledge transfer to less resourced languages. Note that not all language pairs have sufficient data sets for test, development ($dev$) and training ($train$) data. Hence, we divided the Tatoeba challenge data into various subsets based on the size of the training data available.

\begin{description}
    \item[high-resource settings:] 298 language pairs with training data of at least one million training examples (aligned sentence pairs), we further split into language pairs with more than 10 million training examples (173 language pairs) and other language pairs with data sets below the size of 10 million examples
    \item[medium-sized resource settings: ] 97 language pairs with more than 100,000 and less than 1 million training examples
    \item[low-resource settings: ] 87 language pairs with less than 100,000 training examples, we further distinguish between language pairs with more than 10,000 training examples (63) and language pairs below 10,000 training examples (24)
    \item[zero-shot translation: ] language pairs with no training data (40 in the current data set)
\end{description}

\begin{figure*}
    \centering
    \includegraphics[height=4.6cm]{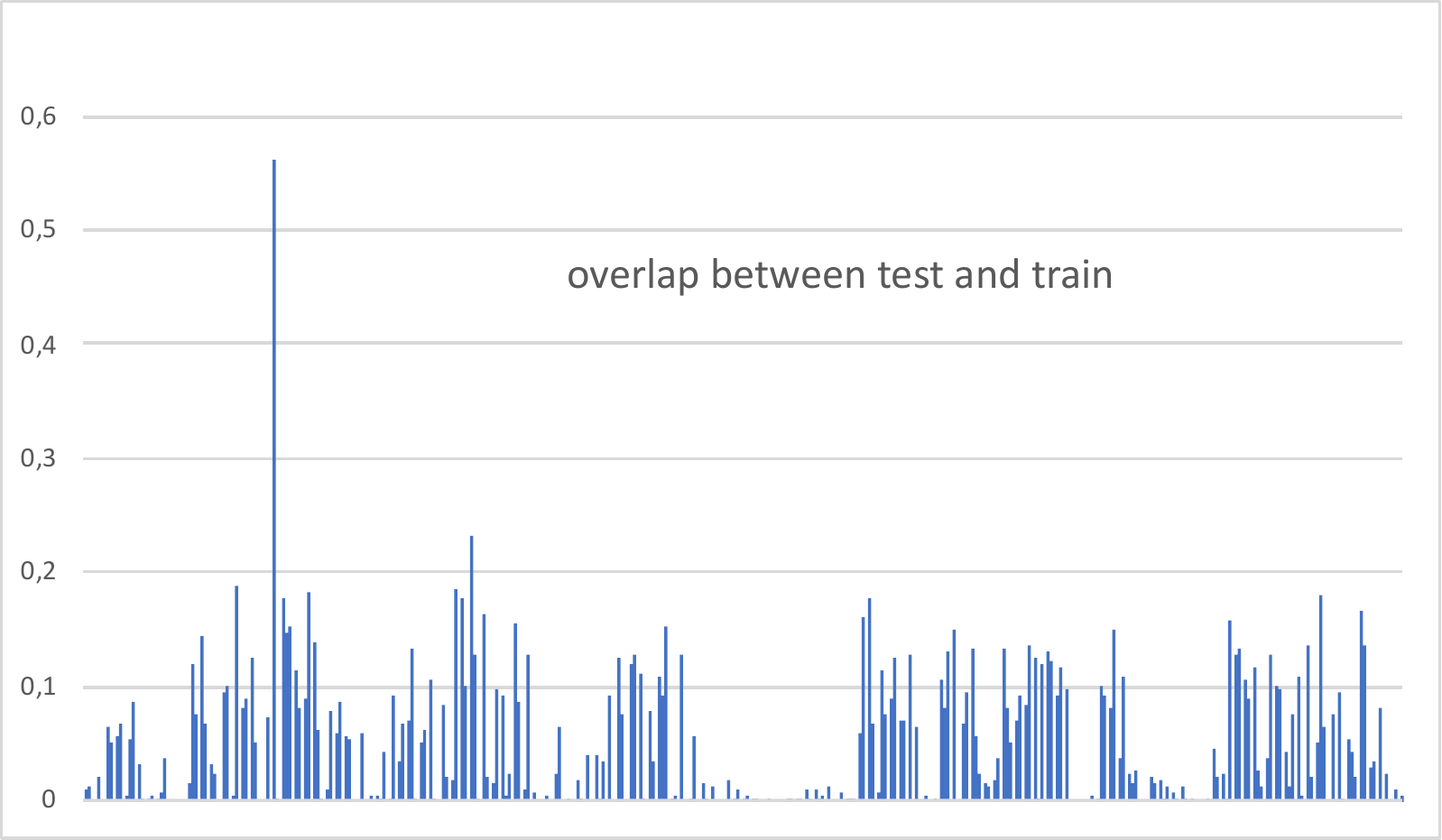}
    \hfill
    \includegraphics[height=4.6cm]{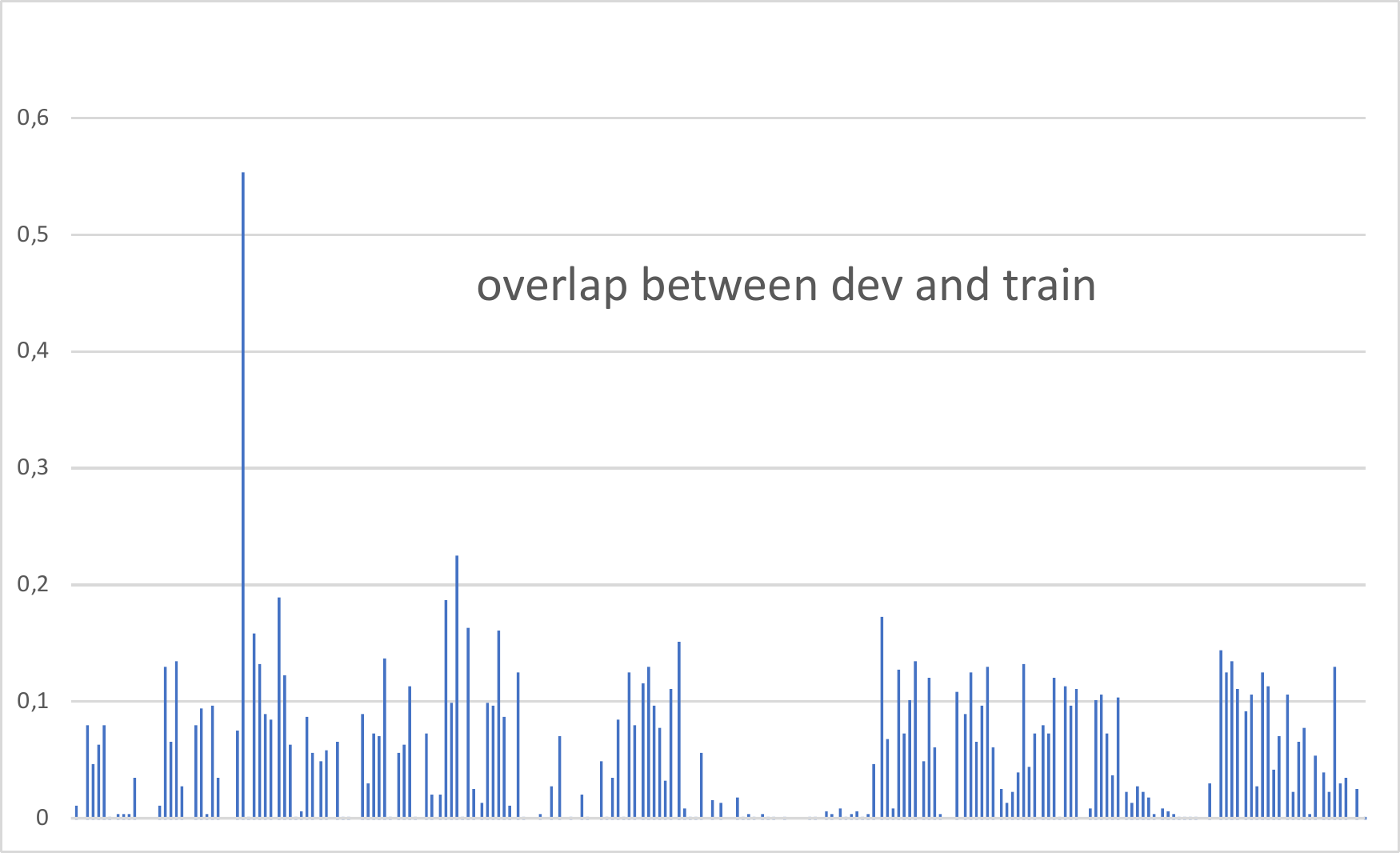}
    \caption{Overlap between test and validation (dev) data and the training data: Proportion of sentence pairs that exist in the training data for all data sets above 1,000 sentence pairs.}
    \label{fig:overlap}
\end{figure*}

For all those 522 selected language pairs, the data set provides at least 200 sentences per test set. 101 of them involves English as one of the languages. 288 test sets contain more than 1,000 sentence pairs of which only 68 include English. Note, that everything below 1,000 sentences is probably not very reliable as a proper test set but we decided to release smaller test sets as an initial benchmark to trigger further development even for extremely under-resourced language pairs. We also decided to use very low thresholds for the division into low-resource languages. Having 10,000 training examples or less is very realistic for many real-world examples and we want to encourage the work on such cases in particular.

The maximum size of test sets in our collection is 10,000 sentence pairs, which is available for 76 language pairs. The test size is reduced to 5,000 if there is less than 20,000 sentence pairs in Tatoeba (19 data sets). The remaining sentences are released as disjoint validation data. For 48 Tatoeba language pairs with less than 10,000 sentence pairs, we keep 2,500 for the test set and the rest for validation and for 78 Tateoba language pairs with less than 5,000 sentence pairs we keep 1,000 for validation and the rest for testing. Finally, for language pairs with less than 2,000 sentences in Tatoeba we skip validation data and use everything for test purposes.

Test and validation data are strictly disjoint and none of the examples from Tatoeba are explicitly included in the training data. However, as it is common in realistic cases, there is a natural chance for a certain overlap between those data sets. Figure~\ref{fig:overlap} plots the percentage of sentence pairs in test and validation sets that can also be found in the corresponding training data we release. The average proportion is rather low around 5.5\% for both with a median percentage of 2.3\% and 2.9\% for test and validation data, respectively. There is one clear outlier with a very high proportion of over 55\% overlap and that is Danish--English for some reason that is not entirely clear to us. Otherwise, the values are well below that ratio.


\section{The data challenge}

The most important ingredient for improved translation quality is data. It is not only about training data but very much also about appropriate test data that can help to push the development of transfer models and other ideas of handling low-resource settings. Therefore, another challenge we want to open here is the increase of the coverage of test sets for low-resource languages. Our strategy is to organize the extension of the benchmarks directly through the Tatoeba initiative. Users who would like to contribute to further MT benchmark development are asked to register for the open service provided by Tatoeba and to upload new translations in the languages of interest. From our side, we will continuously update our challenge data set to include the latest data releases coming from Tatoeba including new language pairs and extended data sets for existing language pairs. We will make sure that the new test sets do not overlap with any released development data from previous revisions to enable fair comparisons of old models with new benchmarks. The extended test and validation data sets will be released as new packages and old revisions will be kept for replicability of existing scores.

In order to provide information about language pairs in need, we provide a list of data sets with less than 1,000 examples per language pair. In the current release, this refers to 2,375 language pairs. 2,141 language pairs have less than 200 translation units and are, therefore, not included in the released benchmark test set. Furthermore, we also provide a list of languages for which we release training data coupled with English but no test data is available from Tatoeba. Currently, this relates to 246 languages.

We encourage users to especially contribute translations for those data sets in order to improve the language coverage even further. We hope to trigger a grass-root development that can significantly boost the availability of development and test sets as one of the crucial elements for pushing NMT development in the corresponding languages.

Finally, we also encourage to incorporate other test sets besides of the Tatoeba data. Currently, we also test with WMT news test sets for the language pairs that are covered by the released development and test sets over the years of the news translation campaign. Contributions and links can be provided through the repository management interface at github.

\section{How to participate}

The goal of the data release is to enable a straightforward setup for machine translation development. Everyone interested is free to use the data for their own development. A leader board for individual language pairs will be maintained. Furthermore, we also intend to make models available that are listed in the challenge. This does not only support replicability but also provides a new unique resource of pre-trained models that can be integrated in real-world applications or can be used in further research, unrelated downstream tasks or as a starting point for subsequent fine-tuning and domain adaptation. A large number of models is already available from our side providing baselines for a large portion of the data set. More details will be provided below.

For participation, there are certain rules that apply:

\begin{itemize}
    \item Do not use any development or test data for training ($dev$ can be used for validation during training as an early stopping criterion).
    \item Only use the provided training data for training models with comparable results in constrained settings. Any combination of language pairs is fine or backtranslation of sentences included in training data for any language pair is allowed, too. That means that additional data sets, parallel or monolingual, are not allowed for official models to be compared with others. \item Unconstrained models may also be trained and can be reported as a separate category.
    Using pre-trained language or translation models fall into the unconstrained category. Make sure that the pre-trained model does not include Tatoeba data that we reserve for testing. 
    \item We encourage to release models openly to ensure replicability and re-use of pre-trained models. If you want to enter the official leader board you have to make your model available including instructions on how to use them.
\end{itemize}

\section{Baseline Models}

Along with the data, we also release baseline models that we train with state-of-the-art transformer models \cite{DBLP:journals/corr/VaswaniSPUJGKP17} using Marian-NMT,\footnote{https://marian-nmt.github.io} a stable production-ready NMT toolbox with efficient training and decoding capabilities \cite{junczys-dowmunt-marian:-2018}. We apply a common setup with 6 self-attentive layers in both, the encoder and decoder network using 8 attention heads in each layer. The hyper-parameters follow the general recommendations given in the documentation of the software.\footnote{https://github.com/marian-nmt/marian-examples/tree/master/transformer} The training procedures follow the strategy implemented in OPUS-MT \cite{TiedemannThottingal:EAMT2020} and detailed instructions are available from github.\footnote{https://github.com/Helsinki-NLP/OPUS-MT-train/blob/master/doc/TatoebaChallenge.md}

We train a selection of models on v100 GPUs with early-stopping after 10 iterations of dropping validation perplexities. We use SentencePiece \cite{kudo-richardson-2018-sentencepiece} for the segmentation into subword units and apply a shared vocabulary of a maximum of 65,000 items. Language label tokens in the spirit of \citet{johnson2017google} are used in case of multiple language variants or scripts in the target language. Models for over 400 language pairs are currently available and we refer the reader to the website with the latest results. For illustration, we provide some example scores below in Table~\ref{tab:medium-eng-results} using automatic evaluation based on chrF2 and BLEU computed using sacrebleu \cite{post-2018-call}. The actual translations are also available for each model and the distribution comes along with the logfiles from the training process and all necessary data files such as the SentencePiece models and vocabularies.

\begin{table}[ht]
    \centering
    \begin{tabular}{c|cc}
    language pair & chrF2 & BLEU \\
    \hline
    aze-eng  &  0.490  &  31.9\\
    bel-eng  &  0.268  &  10.0\\
    cat-eng  &  0.668  &  50.2\\
    eng-epo  &  0.577  &  35.6\\
    eng-glg  &  0.593  &  37.8\\
    eng-hye  &  0.404  &  16.6\\
    eng-ilo  &  0.569  &  30.8\\
    eng-run  &  0.436  &  10.4\\
    \end{tabular}
    \caption{Translations scores from baseline models trained for a selection of medium-size language pairs (according to our classification) tested on the provided Tatoeba benchmark. We show here models that include English and score above 10 BLEU.}
    \label{tab:medium-eng-results}
\end{table}

\section{Multilingual Models}

One of the most interesting questions is the ability of multilingual models to push the performance of low-resource machine translation. The Tatoeba translation challenge provides a perfect testbed for systematic studies on the effect of transfer learning across various subsets of language pairs. We already started various experiments with a number of multilingual translation models that we evaluate on the given benchmarks. In our current work, we focus on models that include languages in established groups and for that we facilitate the ISO-639-5 standard. This standard defines a hierarchy of language groups and we map our data sets accordingly to start new models that cover those sets. As an example, we look at the task of Belorussian-English translation that has been included in the previous section as well. Table~\ref{tab:multi-beleng-results} summarizes the results of our current models sorted by chrF2 scores.

\begin{table}[ht]
    \centering
    \begin{tabular}{l|cc}
    model & chr-F2 & BLEU \\
    \hline
sla-eng/opus4m  &  0.610  &  42.7\\
sla-eng/opus2m  &  0.609  &  42.5\\
sla-eng/opus1m  &  0.599  &  41.7\\
ine-eng/opus2m  &  0.597  &  42.2\\
ine-eng/opus4m  &  0.597  &  41.7\\
ine-eng/opus1m  &  0.588  &  41.0\\
zle-eng/opus4m  &  0.573  &  38.7\\
zle-eng/opus2m  &  0.569  &  38.3\\
mul-eng/opus1m  &  0.550  &  37.0\\
mul-eng/opus2m  &  0.549  &  36.8\\
zle-eng/opus1m  &  0.543  &  35.4\\
ine-ine/opus1m  &  0.512  &  31.8\\
\hline
bel-eng/opus  &  0.268  &  10.0\\
    \end{tabular}
    \caption{Translation results of the Belorussian-English test set using various multilingual translation models compared to the baseline bilingual model (shown at the bottom). opusXm refers to sampled data sets that include X million sentences per language pair.}
    \label{tab:multi-beleng-results}
\end{table}

The models focus on different levels of relatedness of the languages and range from East Slavic Languages (zle), Slavic languages (sla) to the language family of Indo-European languages (ine) and the set that contains all languages (mul). Each model is trained on sampled data set in order to balance between different languages. The smallest training sets are based on data that are sampled to include a maximum of one million sentence per language pair (opus1m). We use both, down-sampling and up-sampling. The latter is done by simply multiplying the existing data until the threshold is reached. We also set a threshold of 50 for the maximum of repeating the same data in order to avoid over-representing small noisy data. The one-million models are trained first and form the basis of larger models. We continue training with data sets sampled to two million before increasing to four million sentence pairs. 

The Table shows some interesting patterns. First of all, we can clearly see a big push in performance when adding related languages to the training data. This is certainly expected especially in the case of Belorussian that is closely related to higher-resource-languages such as Russian and Ukrainian. Interesting is that the East Slavic language group is not the best performing model even though it includes those two related languages. The additional information from other Slavic languages pushes the performance beyond their level quite significantly. Certainly, those models will see more data and this may cause the difference. The 'sla-eng' model covers 13 source languages whereas 'zle' only 5. Also interesting to see is that the Indo-European language model fairs quite well despite the enormous language coverage that this model has to cope with. On the other hand, the big 'mul' translation model does not manage to create the same performance and the limits of the standard model with such a massive setup become apparent. Training those models becomes also extremely expensive and slow and we did not manage to start the 4-million-sentence model.

Currently, we look into the various models we train and many other interesting patterns can be seen. We will leave a careful analyses to future work and also encourage the community to explore this field further using the given collection and benchmark. Updates about models and scores will be published on the website and we would also encourage more qualitative studies that we were not able to do yet.

\section{Zero-shot and few-shot translation}

Finally, we have a quick look at zero-shot and few-shot translation tasks. Table~\ref{tab:zero-shot-results} shows results for Awadhi-English translation, one of the test sets for which no training data is available. Awadhi is an Eastern Hindi language in the Indo-Iranian branch of the Indo-European language family.\footnote{We use ISO639-3 and ISO639-5 standards for names and codes of languages and language groups.}

\begin{table}[ht]
    \centering
    \begin{tabular}{l|cc}
    model & chr-F2 & BLEU \\
    \hline
ine-eng/opus1m  &  0.285  &  10.0\\
mul-eng/opus1m  &  0.257  &  9.4\\
inc-eng/opus1m  &  0.217  &  6.8\\
iir-eng/opus1m  &  0.214  &  7.9\\
ine-ine/opus1m  &  0.201  &  2.4\\
\hline
tatoeba-zero/opus  &  0.042  &  0.1\\
    \end{tabular}
    \caption{Translation results of the Awadhi-English test set using multilingual translation models.}
    \label{tab:zero-shot-results}
\end{table}

The table shows that a naive approach of throwing all languages that are part of zero-shot language pairs into one global multilingual model (tatoeba-zero) does not work well. This is probably not very surprising. Another interesting observation is that a symmetric multilingual model with Indo-European languages on both sides (ine-ine) also underperforms compared to other multilingual models that only translate into English. Once again, the Indo-European-language-family to English model performs quite well. 
Note that the performance purely comes from overlaps with related languages as no Awadhi language data is available during training. The performance is still very poor and needs to be taken with a grain of salt. They demonstrate, however, the challenges one faces with realistic cases of zero-shot translation.

In Table~\ref{tab:few-shot-results}, we illustrate another case that could be described as a realistic few-shot translation task. Our collection comes with 3,613 training examples for the translation between English and Faroese. The table shows our current results in this task using multilingual models that translate from English to language groups including the Scandinavian language in question. 

\begin{table}[ht]
    \centering
    \begin{tabular}{l|cc}
    model & chr-F2 & BLEU \\
    \hline
eng-gem/opus  &  0.318  &  9.4\\
gem-gem/opus  &  0.312  &  7.0\\
eng-gmq/opus  &  0.311  &  7.0\\
eng-ine/opus  &  0.281  &  6.3\\
eng-mul/opus  &  0.280  &  5.7\\
ine-ine/opus  &  0.276  &  5.9\\
\hline
tatoeba-zero/opus  &  0.042  &  0.1\\
    \end{tabular}
    \caption{Translation results of the English-Faroese test set with different multilingual NMT models.}
    \label{tab:few-shot-results}
\end{table}

Again, we can see that the naive tatoeba-zero model is the worst. The symmetric Indo-European model performs better but the English-Germanic model gives the best performance, which is still very low and not satisfactory for real-world applications.
Once again, the example demonstrates the challenge that is posed by extremely low-resource scenarios and we hope that the data set we provide will trigger additional fascinating studies on a large variety of interesting cases.

\section{Comparison to the WMT news task}

Finally, we also include a quick comparison to the WMT news translation task, see Table~\ref{tab:WMT-results}. Note that we did not perform any optimization for that task, did not use any in-domain back-translations and did not run fine-tuning in the news domain. We only give results for English--German (in both directions) for the 2019 test data to give an impression about the released baseline models. 

\begin{table}[ht]
    \centering
    \begin{tabular}{l|cc}
    \hline
    \multicolumn{3}{c}{English -- German} \\
    \hline
    model & BLEU & chr-F2 \\
    \hline
eng-deu & 42.4  & 0.664 \\
eng-gmw & 35.9  & 0.616 \\
eng-gem & 35.0  & 0.613 \\
eng-ine & 26.6  & 0.554 \\
eng-mul & 21.0  & 0.512 \\
\hline
WMT best & 44.9 & -- \\
\hline
    \hline
    \multicolumn{3}{c}{German -- English} \\
    \hline
    model & BLEU & chr-F2 \\
    \hline
deu-eng & 40.5  & 0.645 \\
gmw-eng & 36.6  & 0.615 \\
gem-eng & 37.2  & 0.618 \\
ine-eng & 31.7  & 0.571 \\
mul-eng & 27.0  & 0.529 \\
\hline
WMT best & 42.8 & -- \\
\hline
    \end{tabular}
    \caption{Translation results of baseline models on English--German news translation from WMT 2019 using bilingual and multilingual Tatoeba baseline models. The BLEU scores are also compared to the best score that is currently available from http://matrix.statmt.org/matrix -- retrieved on October 4, 2020.}
    \label{tab:WMT-results}
\end{table}

The results demonstrate that the models can achieve high quality even on a domain they are not optimized for. The best scores in the German--English case are close to the top performing model registered for this task even though the comparison is not fair for various reasons. The purpose is anyway not to provide state-of-the-art models for the news translation task but baseline models for the Tatoeba case and in future work we will also explore the use of our models as the basis for systems that can be developed for other benchmarks and applications. In the example we can also see that multilingual models significantly lag behind bilingual ones in high-resource cases. Each increase of the language coverage (except for the move from West Germanic languages (gmw) to Germanic languages (gem) in the German--English case) leads to a drop in performance but note that those multilingual models are not fine-tuned for translating from and to German.

\section{Conclusions}   \label{sec:conclusions}

This paper presents a new comprehensive data set and benchmark for machine translation that covers roughly 3,000 language pairs and over 500 languages and language variants. We provide training and test data that can be used to explore realistic low-resource scenarios and zero-shot machine translation. The data set is carefully annotated with standardized language labels including variations in scripts and with information about the original source. We also release baseline models and results and encourage the community to contribute to the data set and machine translation development. All tools for data preparation and training bilingual as well as multilingual translation models are provided as open source packages on github. We are looking forward to new models, extended test sets and a better coverage of the World's languages.

\section*{Acknowledgements}

\begin{wrapfigure}{l}{1cm}
\includegraphics[width=1cm]{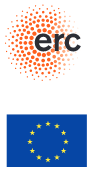}
\end{wrapfigure}
This work is supported by the FoTran project (grant agreement No 771113), funded by the European Research Council (ERC) and the MeMAD project (grant agreement No 780069) under the European Union's Horizon 2020 research and innovation program. We would also like to acknowledge the support of the CSC – IT Center for Science, Finland, for computational resources.

\bibliographystyle{acl_natbib}
\bibliography{tatoeba}

\end{document}